\documentclass{article}
\usepackage[utf8]{inputenc}
\usepackage{amsmath}
\usepackage{amssymb}
\usepackage{graphicx}
\graphicspath{{prototype/results/}{prototype/results_real/}}
\usepackage{booktabs}
\usepackage[hidelinks]{hyperref}
\usepackage{cite}
\usepackage[margin=1in]{geometry}
\usepackage{array}
\usepackage{url}
\hypersetup{
  pdftitle={Entropy in Conversational AI: Structured Unpredictability as Inferrable Interiority},
  pdfauthor={Sebastian Cochinescu},
  pdfsubject={Stateful response selection and prospectively specified evaluation in conversational AI},
  pdfkeywords={conversational AI, decoding, response selection, persona consistency, anthropomorphism}
}

\title{Entropy in Conversational AI: Structured Unpredictability as Inferrable Interiority\\[0.5em]
{\large\normalfont Identity as Accumulated Selection: A Formal Model, a Mechanism, and a Prospectively Frozen Divergence Test}}

\author{
Sebastian Cochinescu\\
\textit{University of Bucharest}\\
\textit{Email: sebastian.cochinescu@drd.unibuc.ro}
}

\date{July 2026}

\begin{document}

\maketitle

\begin{abstract}
Sampling can increase response diversity without producing history-dependent behavior. We formalize a different design target, \emph{structured unpredictability}, as conditional dependence between an output and a persistent hidden state beyond what an observer can infer from the transcript. A selection layer updates a low-dimensional style-and-attention state from a capacity-limited stream, generates several responses with a fixed base model, and selects for novelty and state affinity. Evaluation uses scripted sequences of independent prompt turns: the base model receives the current turn and rendered state, but not the preceding dialogue; cross-turn dependence resides in the wrapper state and response selector. A synthetic implementation validates the pipeline and matches four prospectively hash-frozen divergence features at point level. In the final real-model grid (\texttt{mlx-community/Qwen2.5-1.5B-Instruct-4bit}; 56 sequences per arm), the mechanism increased lexical novelty over the low-variance and consistency-only controls by $0.073$ and $0.023$, respectively. Its stylometric-consistency contrast with novelty-matched sampling was equivalent to zero under the registered smallest-effect rule, so the joint novelty--consistency criterion failed. The original two-part accumulation criterion also failed; a revised final-grid contrast, frozen after the powered grid, found higher consistency than the memory-reset ablation ($0.028$, 95\% CI $[0.018,0.039]$), but does not establish path dependence. Twin separation was not established ($0.003$, 95\% CI $[-0.011,0.019]$); the mean curve's saturating curvature matched the frozen prediction, which without separation does not support path dependence. Probe-level capability equivalence held within $\pm0.10$ on a near-ceiling battery, while output quality was not evaluated. All outcomes are machine-scored; no claims about perceived mind or consciousness are tested.
\end{abstract}


\section{The Problem: Noise Is Not Depth}
\label{sec:problem}

Conversational systems often obtain variation by changing decoding parameters. That variation need not reflect any state acquired through interaction. The series' framework paper calls the relevant stance \emph{entropy}: structured, identity-consistent variation over time, distinct from output diversity alone~\cite{cochinescu2026perceivedagi}. This paper asks whether that stance can be specified as a stateful selection mechanism and evaluated without attributing mental states to the system. Because apparent interiority can affect attachment, the mechanism and its negative results should be inspectable and tested against fixed criteria (Section~\ref{sec:series}).

Likelihood-based decoding exhibits two familiar failure modes: maximization can produce bland or repetitive text, while less restrictive sampling can increase variation without preserving a stable pattern~\cite{holtzman2020curious}. Temperature, top-$k$, and nucleus mass move generation along a quality--diversity trade-off~\cite{fan2018hierarchical,caccia2020language}; repetition is itself a controllable dialogue attribute with measurable effects on human judgments~\cite{see2019whatmakes}. When these decoders are used without retained interaction state, as in the sampling controls evaluated here, each turn is generated without conversational history. Increased temperature then changes sampling variance, not cross-turn state.

The proposed alternative is observer-relative: outputs remain difficult to predict from the transcript alone but become more predictable when conditioned on a persistent hidden state. This is the paper's operational meaning of structured unpredictability. Whether people interpret that pattern as character or mind is a separate empirical question that is not tested here.

The contribution is narrower than a general account of adaptive identity. We define \emph{identity as accumulated selection}: a low-dimensional state changes as the wrapper selects from a capacity-limited stream and from current-turn features; that state then conditions candidate generation and response selection. The model yields an observer-conditional entropy gap and a prospectively frozen twin-divergence prediction (Section~\ref{sec:model}). The mechanism is evaluated in seven arms over a fixed base model (Sections~\ref{sec:mechanism}--\ref{sec:protocol}). The evidence consists of synthetic pipeline validation and staged real-model experiments (Sections~\ref{sec:results}--\ref{sec:results:final}). Relative to the closest prior systems, authored-persona and trained-character approaches fix the identity target before interaction, and decoding-time steering or best-of-$n$ reranking carries no persistent target at all; here the selection target is constituted by the interaction history itself, and that construction is converted into a prospectively frozen quantitative prediction rather than a post-hoc consistency score. On the tested real model most registered criteria fail; the durable contributions are the formal target, the frozen twin-divergence methodology, and the registered negative results, which later work can rerun against larger or more style-pliable base models. All measurements are machine-side; the proposed human-perception effect remains for a separate study.

\section{The Model: Identity as Accumulated Selection}
\label{sec:model}

The model is a design specification, not a cognitive theory: its role is to make ``structured unpredictability'' a definable property and to emit a falsifiable quantitative prediction, not to describe how identity works in humans.

\subsection{State, stream, selection, drift}
\label{sec:model:core}

An agent carries a persistent \textbf{hidden state} $H_t \in \mathbb{R}^K$. Its coordinates are operational selection parameters---style weights and topical attention weights. Where we gloss coordinates with psychological words (a ``mood,'' a ``value,'' an ``attention bias''), the gloss is interpretive only; nothing in the mathematics depends on it.

At each step the environment offers an \textbf{ambient stream} $E_t$: a set of $m$ candidate experience items, represented in the same feature space as $H_t$, of which the agent can attend to at most $b < m$. The strict inequality is necessary for selective exposure. Under the model, differences in admitted items can accumulate even when agents begin from the same state; ``identity'' refers only to that selection-history construct.

A \textbf{selection function} makes the choice: $s_t = \sigma(H_t,\, E_t \cup \{\text{turn}_t\})$ scores the stream items together with the current conversational turn and admits at most $b$, combining two pressures whose joint optimization is the mechanism's signature: \emph{identity pressure} (affinity between an item and the current state, computed on deviations from a neutral point---a fixed constant of the featurization, style coordinates at $0.5$ and topics at the uniform mixture---so affinity is rich-get-richer and selection genuinely path-dependent) and \emph{novelty pressure} (a penalty on similarity to recently selected experience).

The state then \textbf{drifts}:
\begin{equation}
H_{t+1} \;=\; f\!\left(H_t,\; s_t,\; O_t,\; \varepsilon_t\right)
\;=\; \operatorname{clip}\!\left[(1-\lambda)\,H_t \;+\; \lambda\,\overline{s_t} \;+\; \gamma\, O_t\, u \;+\; \varepsilon_t\right],
\label{eq:drift}
\end{equation}
an exponentially weighted accumulation of selected experience $\overline{s_t}$, perturbed by small noise $\varepsilon_t$ and modulated by an \textbf{internal oscillation term} $O_t$, imported from the series' time paper's Van der Pol architecture for endogenous temporal state~\cite{cochinescu2025scnllm} and entering here as a fixed-parameter oscillator coupled to the style coordinates through a unit pattern $u$. Concretely, $O_t$ is the position of a Van der Pol unit $\ddot{x} = \mu(1-x^{2})\dot{x} - \omega^{2}x$ with $\mu=0.5$ and $\omega=2\pi/40$ (one cycle per forty turns), integrated open-loop by fixed-step RK4 from a seeded uniform-random initial phase and normalized to $[-1,1]$. Two properties matter for the divergence experiment. It is \emph{open-loop}---evolving autonomously, decoupled from $H_t$ and $\varepsilon_t$, so agents sharing an initialization share its phase exactly---and therefore common-mode: an observer can fit a fixed-parameter periodic signal from a long enough transcript, so asymptotically it contributes no observer-side unpredictability (over finite prefixes it can), and at every horizon it contributes no identity differentiation and no residual entropy-gap term in the sense of Section~\ref{sec:model:gap}. Its role is temporal texture and series continuity; the drift model is well-defined with it ablated, and none of the evaluation contrasts depend on it.

\subsection{Structured unpredictability, formalized}
\label{sec:model:gap}

Let $(H_t, T_t, Y_t)$ be the hidden state, the transcript prefix, and the next response, jointly distributed under the mechanism's stochastic dynamics (stream, selection, noise, decoding). Define
\begin{equation}
\Delta_t \;=\; \mathcal{H}(Y_t \mid T_t) \;-\; \mathcal{H}(Y_t \mid T_t,\, H_t) \;=\; I(Y_t;\, H_t \mid T_t).
\label{eq:gap}
\end{equation}
As a conditional mutual information~\cite{cover2006elements}, $\Delta_t \ge 0$, with equality exactly when the response is conditionally independent of the state given the transcript. Strict positivity, the design target, holds when the state carries information about the response beyond what the transcript reveals. A stateless randomizer has $\Delta_t=0$ under this definition even when its marginal output entropy is high; a deterministic script has little marginal output entropy to explain. The proposed target is high marginal variation together with nonzero conditional dependence on $H_t$. Positivity alone also does not distinguish accumulated state from fixed hidden conditioning---a static persona vector or a per-agent seed satisfies it equally; separating accumulation from fixed conditioning is the twin experiment's burden (Section~\ref{sec:model:twins}). No human interpretation follows from this quantity alone.

We are explicit about what this definition does and does not do. $\Delta_t > 0$ is a design target that the mechanism of Section~\ref{sec:mechanism} is built to satisfy: the construction supplies the causal path (the state exists, drifts, and conditions generation) but not a guarantee---the response could still be conditionally independent of $H_t$, the equality case above, and the pilot's transmission failure (Section~\ref{sec:results:pilot}) is exactly the case where the path carries nothing---so positivity is a property to be earned per model, never assumed from the architecture. It is also \emph{not} measured here: conditional-entropy estimation for text is estimator-dependent, and no claim rests on such an estimate. The measured axes of Section~\ref{sec:model:axes} are observable proxies motivated by the gap's two regimes---they are not its mathematical sides, neither estimates a conditional entropy, and the evaluation never treats them as if they did.

Accordingly, \emph{entropy} here denotes structured behavioral variation, not token-level surprisal, consistent with the framework paper's definition~\cite{cochinescu2026perceivedagi}.

\subsection{The two axes}
\label{sec:model:axes}

The gap's two regimes suggest two positively oriented, $[0,1]$-valued observables, computed always against the agent's \emph{own} history, never against an authored script.

The \textbf{novelty axis} $N_t$ operationalizes observer-side unpredictability as non-repetition: one minus the fraction of the current turn's word bigrams already present in the agent's pooled prior outputs, length-normalized---the Self-BLEU/distinct-$n$ family~\cite{zhu2018texygen,li2016diversity}. A pre-specified semantic variant (topic-vector similarity against recent turns) is computed alongside; under recurring-intent scripts the semantic channel sits near zero by design (topic recurrence is the controlled repetition opportunity), so the operative test is lexical and the semantic variant serves as a guard, not a second effect channel.

The \textbf{consistency axis} $C_t$ operationalizes coherence-given-state as the stability of the agent's outputs around its own accumulated \emph{stylometric} profile: function-word distributions, sentence-length statistics, and type--token ratio---authorship-verification-class features~\cite{stamatatos2009survey}. The representation choice is a designed defense against the confound that would otherwise collapse the two axes: semantic embeddings track content, mechanically coupling consistency to novelty, whereas content-independent style representations exist precisely to separate the two~\cite{wegmann2022same}. A semantic-embedding variant of $C_t$ is reported only as a secondary robustness check.

The \textbf{entropy stance} is the joint property: high $N_t$ \emph{and} high $C_t$, sustained as $t$ grows---the high/high quadrant of a plane whose axes the sampling literature usually treats as a trade-off. Because the axes could fail to be two axes at all, the protocol builds in validity controls rather than asserting separation: perturbation arms that move one axis at a time, a pure-sampling grid tracing the knob frontier, a topic-permutation check (stylometric $C_t$ recomputed over topic-permuted transcripts must not move if it measures voice rather than content), a between-agent distinctiveness test (an agent's turn should match its own history better than another agent's), and a deliberately bland control (coherent boilerplate must expose, through the joint criterion, why stability alone is not voice).

\subsection{The twin-divergence prediction}
\label{sec:model:twins}

The model makes a quantitative prediction before text generation is measured. We simulate Eq.~\eqref{eq:drift} directly in state space---without a language model---for two agents with identical $H_0$ and oscillator phase. \emph{Different histories} use different stream and turn sequences; \emph{same history} uses the same sequences with independent $\varepsilon_t$ and defines the comparison floor. The simulation produces a divergence-versus-horizon band ($\ell_2$ state distance) over seeded pairs.

To prevent arbitrary rescaling from determining agreement, the comparison uses four dimensionless features computed on raw curves: monotonicity; curvature class (saturating, linear, or accelerating); the half-to-full-horizon ratio after baseline subtraction; and the ratio of median final divergence under different versus same history. The last feature is discriminating: a static persona or per-turn reranker predicts no above-floor history effect. The remaining features are compatible with many accumulating processes, so agreement would support only path-dependent divergence of the specified shape, not the particular function $f$.

The prediction is generated first and frozen---file hash recorded, regeneration byte-identical from the seed, the reproduction script refusing to run if the file drifts---before any generation run. For visualization only, the predicted band is overlaid on measurements through a strictly affine map (scale positive, no warping) fit on pilot pairs and applied once; confirmatory shape features are computed on raw curves.

\section{Mechanism: A Selection Layer over a Fixed Base Model}
\label{sec:mechanism}

The mechanism instantiates the model on top of an unmodified base language model, continuing the series' pattern of realizing a stance as a layer over a fixed model rather than as a capability change~\cite{cochinescu2025scnllm}. Nothing is trained; the base model is identical in every experimental arm.

Each turn proceeds in four steps. (1)~The ambient-stream simulator emits $m$ candidate experience items; the current turn's features join the pool. Items come from a fixed world of topic clusters shared by every agent and stream seed: each item carries a Dirichlet-concentrated topic mixture around its cluster and the cluster's style center perturbed by small noise; the real runs use $m{=}12$ items per turn and budget $b{=}3$. (2)~The selection function admits $b$ items by the joint objective---identity affinity on centered features minus a damped novelty penalty on similarity to recently selected experience---and the state drifts by Eq.~\eqref{eq:drift}. (3)~The base model generates $n$ candidate responses conditioned on a mixture of the state and the turn. (4)~The response selector picks the candidate maximizing the joint objective at the output level: novelty pressure against the agent's own past outputs at full weight, plus identity affinity between the candidate's features and $H_t$. The asymmetry---novelty damped on experience selection, primary on response selection---is recorded in the protocol: attention is where identity locks in, output where repetition is policed.

Decoding-time control methods steer generation toward an attribute by gradient steering~\cite{dathathri2020pplm}, discriminator guidance~\cite{yang2021fudge,krause2021gedi}, contrastive objectives~\cite{li2023contrastive}, or control codes and repetition penalties~\cite{keskar2019ctrl}. Best-of-$n$ selection scores candidates against a criterion~\cite{stiennon2020learning}; persona-grounded dialogue supplies an authored profile~\cite{zhang2018personalizing}; trained role-play agents encode character behavior during training~\cite{shao2023characterllm}. The memory-reset and single-objective arms instantiate close controls for these static or per-turn alternatives.

Interaction-dependent systems make a categorical novelty claim untenable. Generative agents use episodic memory, reflection, and planning~\cite{park2023generative}; PersonaAgent revises a user-specific persona prompt from interaction-derived memory~\cite{zhang2025personaagent}; and TMEM changes an agent's within-episode policy through online low-rank parameter updates~\cite{ren2026tmem}. These systems target memory, personalization, or task adaptation. The present mechanism instead retains a low-dimensional control state with no retrievable episodes and leaves model weights fixed. Its distinctive evaluation target is whether capacity-limited selection produces the pre-specified novelty--consistency contrasts and above-floor twin divergence. This is a difference in mechanism and estimand, not a claim that earlier systems cannot adapt.

Persona-stability work measures a complementary phenomenon: departure from an authored persona is treated as failure. Instruction-stability experiments study shorter dialogues~\cite{li2024instability}; agent--agent conversation exhibits \emph{echoing}, agents abandoning assigned roles to mirror their partners~\cite{shekkizhar2025echoing}; SPASM reduces role and persona drift in generated multi-turn dialogue~\cite{luo2026spasm}; and ContextEcho audits drift across thousands of turns in agentic coding sessions~\cite{ding2026contextecho}. Here, movement induced by selected history is the hypothesized signal, but the real-model experiment does not reproduce those long-context settings: it uses independent prompt turns to isolate the wrapper. The twin prediction is related to trajectory analysis in nonlinear recurrent systems~\cite{sompolinsky1988chaos,sussillo2013opening}, but the test concerns four pre-specified curve features rather than identification of the particular drift function.

\section{Evaluation Protocol}
\label{sec:protocol}

The protocol proceeds from a synthetic stand-in to a real-model pilot, a renderer manipulation check, a powered grid, and a final grid (Sections~\ref{sec:results}--\ref{sec:results:final}). Constants and decision rules are versioned as frozen, provisional, or pending in the artifact protocol. The divergence prediction was hash-frozen before any text-generation run. The final-grid design and analysis were committed before execution; the OSF registration (\url{https://osf.io/dyt3w}) was created after the registered calibration and matching steps but before confirmatory outputs (Section~\ref{sec:results:final}). We therefore distinguish the prospective repository freeze from the later independent-registry timestamp; the freeze-and-registry discipline is adapted from registered-report methodology~\cite{chambers2022registered}.

\subsection{Arms}
\label{sec:protocol:arms}

Seven arms, all sharing one base model and one seeded pipeline, differing only in configuration:

\begin{enumerate}
\item \textbf{mechanism} --- the full selection layer (novelty $+$ identity pressure, persistent state, ambient stream);
\item \textbf{low-variance control} --- greedy/low-temperature pure sampling: the bland floor, a fixed point, not a tunable budget;
\item \textbf{novelty-matched randomness control} --- pure sampling with temperature tuned so its measured $N_t$ matches the mechanism's, without which the consistency contrast would be a tautology at unequal diversity budgets;
\item \textbf{memory-reset ablation} --- the full selection layer with the hidden state returned to $H_0$ every turn; the oscillator phase, the recent selected-experience record, and the selector's own-output novelty history persist, so the ablation removes $H$-accumulation specifically: the stateful, non-accumulating rerank baseline;
\item \textbf{novelty-only selector} --- identity weight zero: the repetition-penalty / single-objective class;
\item \textbf{consistency-only selector} --- novelty weight zero: the static-attribute class;
\item \textbf{pure-sampling grid} --- a temperature $\times$ nucleus-mass sweep tracing the knob frontier that the mechanism is claimed to escape.
\end{enumerate}

The archived implementation fixes the constants: $K=12$ features (six style groups, six topics); $n=4$ candidate responses per turn in the real runs' selection arms (a run configuration; sampling arms emit a single response); drift learning rate $\lambda=0.08$, oscillator coupling $\gamma=0.05$, state-noise scale $0.002$, clipping to $[0,1]^K$; selection weights $1.0$ for both pressures in the mechanism arm, zeroed per single-objective arm; low-variance control at temperature $0.1$, top-$p$ $0.7$; grid sweep over temperatures $\{0.4,0.9,1.5\}$ at nucleus mass $0.9$ plus $(0.9,0.7)$; 12-turn sequences with a 6-turn final window in the real grids, 24 turns for twins. $C_{\mathrm{sty}}$ is one minus the normalized $L^1$ distance between the current turn's stylometric profile and the mean profile of the agent's own history; $N_{\mathrm{lex}}$ is the pooled-bigram non-overlap of Section~\ref{sec:model:axes}.

\subsection{Estimand, matching, and statistics}
\label{sec:protocol:stats}

The statistical unit is a scripted turn sequence (called a ``conversation'' in the archived code and filenames). Each sequence cycles recurring intents to create controlled repetition opportunities. Crucially, every base-model call contains only the current rendered system instruction and current user prompt; the preceding dialogue is not supplied. Cross-turn dependence in selection arms is carried by $H_t$, recent selected-experience features, and the selector's record of its own prior outputs. Sampling controls carry no semantic history. Prompts and seeds are crossed factors, and intervals resample whole sequences to avoid turn-level pseudoreplication. Each arm runs the same fixed script battery under a shared per-script seed schedule, one sequence per script, and arms are paired by script index. Contrast tests use sign flips of within-script paired differences, exact for the sharp null that two configurations are behaviorally equivalent: within a script block the arms share the script and seed schedule, so exchanging arm labels is a symmetry of that null and the paired difference is symmetric about zero. Hypothesis tests are therefore conditional on the fixed battery and seed schedule; bootstrap intervals additionally treat scripts as exchangeable draws, so they describe uncertainty across scripts like those in the battery rather than a wider prompt population. The estimand is the mean of each metric over the fixed final window.

The joint manipulation check (C2) is a conjunction: the absolute $N$ and $C$ thresholds, four directional contrasts with Holm multiplicity control~\cite{holm1979simple}, and a position outside the tested pure-sampling envelope must all pass. The original accumulation criterion (C3) separately required an arm-specific rise in $C_t$ and loss of that rise under memory reset. After the powered grid, the final-grid registration added a different paired contrast, mechanism versus memory reset on $C_{\mathrm{sty}}$. The same registration fixed a smallest effect of interest~\cite{lakens2017equivalence} for consistency, $\mathrm{SME}_C=0.02$---approximately two pilot sequence-level standard deviations, about twice the powered grid's cross-arm spread---declaring a null consistency contrast equivalent to zero when its paired bootstrap 95\% upper bound lies below $\mathrm{SME}_C$. The registered rule is one-sided---it excludes a beneficial consistency gain of $\mathrm{SME}_C$ or more; the observed matched-sampling interval also lies within the symmetric $\pm0.02$ band (Section~\ref{sec:results:final}). We report the original C3 and this later state-retention contrast separately.

Matched temperature is estimated by bisection on a calibration split and then applied to untouched confirmatory seeds; an out-of-bracket target uses the nearest endpoint and is flagged as unmatched. Capability parity is evaluated against prospectively specified $\pm0.10$ equivalence margins---a stipulated one-item-in-ten accuracy band, not an externally derived threshold---on a fixed battery of ten short keyword-scored factual questions crossed with generation seeds. The battery is near ceiling for the grid model, so equivalence within the margin bounds gross accuracy loss only. The protocol also proposed machine-scored output quality, but that component was not run; all capability statements are therefore restricted to the probes and latency.

\subsection{Twin-divergence protocol}
\label{sec:protocol:twins}

Twin pairs share the base model, code, $H_0$, and open-loop oscillator phase. The \emph{different-history} condition uses independent stream and script sequences; the \emph{same-history} condition shares streams, scripts, and generation seeds and differs only in the state-noise seed. Because neither condition supplies dialogue history to the base model, the comparison is between the same closed-loop wrapper dynamics, not between autoregressive conversations. State-dependent candidate generation and output-history-dependent selection can still amplify small state differences. A fixed-candidate diagnostic was reserved for a floor large enough to threaten the comparison.

At fixed checkpoints both twins are probed with shared generation seeds, each probe conditioning on the agent's state alone---rendered, as on every turn, through the fixed voice-instruction template, never as raw numbers---so the probe asks the agent about its own current dominant interest, and whatever the two histories have made different is what carries the signal. The divergence estimator is deliberately not a single sampled response---the generator consumes randomness data-dependently, so any nonzero state difference desynchronizes the random streams, turning a single-response comparison into a near-binary chaos detector rather than a distance. Instead each agent's \emph{probe voice profile} is the mean stylometric profile over a batch of candidates generated from its conditioning, and the reported divergence subtracts a split-half estimate of the estimator's own noise floor. The estimator is fully specified: per probe (three per checkpoint, batches of twelve candidates per agent in the real runs), the coordinatewise absolute difference of the two agents' mean profiles, minus the agents' averaged within-batch split-half difference divided by $\sqrt{2}$ (the scaling matching the null expectation of the between-agent difference), clipped at zero, averaged over coordinates and probes. The $\sqrt{2}$ scaling assumes independent, equal-variance half-batches under the null; shared seeds correlate the agents' batches, shrinking the raw difference---erring conservative---and the paired twin analysis differences two identically corrected conditions, cancelling most residual bias. The clip is what produces exact zeros at the floor; the estimator is marked provisional in the protocol, and a simulation-based calibration archived with the final grid's sizing (Gaussian profiles at batch twelve) measures a null mean ${\approx}0.006$, 95th percentile ${\approx}0.010$, and downward bias under true separation---conservative rather than anti-conservative under the simulated conditions, a model-based calibration that does not certify other coordinate distributions; twin statistics are read within that resolution throughout. Divergence is stylometric-primary with a semantic robustness row, for the same anti-confound reason as $C_t$; probes are excluded from the agents' histories; and the same-history ``floor'' is itself a gently rising curve, not a flat constant---independent state noise accumulates too---so it is reported as a distribution over seed pairs and every floor comparison is an inequality on intervals, never points.

Table~\ref{tab:deviations} consolidates the protocol revisions disclosed at their points of use in Sections~\ref{sec:results}--\ref{sec:results:final}, each with its stage, freeze status, and the direction in which it could bias outcomes. None alters a hash-frozen artifact. Most of the listed revisions are pass-favorable in direction; the bound on this risk is that the anchors---the hash-frozen prediction and the pre-execution repository freezes---were never revised.

\begin{table}[t]
\centering
\small
\begin{tabular}{@{}p{0.29\linewidth}p{0.16\linewidth}p{0.24\linewidth}p{0.22\linewidth}@{}}
\toprule
\textbf{Revision} & \textbf{Stage} & \textbf{Freeze status} & \textbf{Direction} \\
\midrule
Twin-ratio tolerance $\pm0.15 \to \pm0.20$ & synthetic validation & provisional constant; before any real-model run & widens pass region; ratio also passes at $\pm0.15$ \\
Probe conditioning mixed $\to$ state-only & pilot & instrument revision before powered grid & restores muted signal \\
Per-variant calibration thresholds & pilot & specification tightening before powered grid & neutral \\
Inflection-tolerant lexicon matching & renderer gate & instrument revision & raises gate topic scores \\
Ecological topic measurement & renderer gate & instrument revision & aligns gate with pipeline \\
Sign-flip exchangeability group & powered grid & post-data reanalysis, labeled & decisions unchanged \\
Original C3 $\to$ paired reset contrast & after powered grid & frozen before final grid; reported separately & detects effect C3 missed \\
Memory-reset capability TOST & powered grid & post-review addendum; frozen probes, margin, seeds & expands coverage \\
Semantic threshold $\to$ non-inferiority guard & final registration & frozen before final grid & vacuous conjunct becomes guard \\
Twin primary statistic median $\to$ mean & final registration & frozen before final grid & avoids clip zero-inflation \\
\bottomrule
\end{tabular}
\caption{Disclosed protocol revisions across stages. ``Direction'' states how each revision could bias outcomes relative to the original specification.}
\label{tab:deviations}
\end{table}

\section{Results: A Synthetic Reference Implementation}
\label{sec:results}

Everything measured in this section uses a seeded synthetic generator designed to expose low-temperature repetition and high-temperature incoherence. These results validate code paths, metrics, arms, matching, and twin analysis; they are not evidence about an open-weight language model. The ancillary artifact contains the implementation, tests, inputs, and generated results files. Synthetic-run metadata specify master seed 20260711, eight 24-turn sequences per arm, twelve twin pairs per condition over 40 turns, and a prediction band over 48 pairs. The archived powered-run prose summary contains two stale labels documented in \texttt{results\_powered/ERRATA.md}; numeric files are unaffected.

\subsection{Instrument validation}
\label{sec:results:instruments}

The stylometric metric is invariant under topic permutation on mechanism transcripts (maximum shift $0.0$). Between-agent distinctiveness is positive for mechanism agents (self-match minus cross-match $=0.029$) and zero for boilerplate twins. The boilerplate control has $C_{\text{sty}}=1.000$ and $N_{\text{lex}}=0.000$, showing why consistency alone cannot satisfy the joint criterion. Novelty matching converged: the calibration target was $N_{\text{lex}}=0.934$; temperature $2.275$ achieved $0.932$ on calibration and $0.939$ on the confirmatory split (Table~\ref{tab:arms}). Selection overhead was $652\,[590,705]$ microseconds per turn versus $1101\,[999,1220]$ for synthetic generation.

\begin{table}[t]
\centering
\small
\begin{tabular}{@{}lcccc@{}}
\toprule
\textbf{Arm} & $N_{\text{lex}}$ & $N_{\text{sem}}$ & $C_{\text{sty}}$ & $C_{\text{sem}}$ \\
\midrule
mechanism           & 0.931 [0.887, 0.965] & 0.156 [0.114, 0.221] & 0.594 [0.566, 0.679] & 0.752 [0.704, 0.802] \\
low-variance        & 0.509 [0.441, 0.579] & 0.000 [0.000, 0.000] & 0.660 [0.598, 0.714] & 0.587 [0.483, 0.748] \\
randomness-matched  & 0.939 [0.911, 0.974] & 0.126 [0.096, 0.201] & 0.627 [0.542, 0.672] & 0.745 [0.705, 0.828] \\
memory-reset        & 0.998 [0.978, 1.000] & 0.192 [0.117, 0.222] & 0.621 [0.573, 0.643] & 0.731 [0.662, 0.794] \\
novelty-only        & 1.000 [0.994, 1.000] & 0.191 [0.122, 0.276] & 0.579 [0.547, 0.654] & 0.720 [0.635, 0.784] \\
consistency-only    & 0.830 [0.747, 0.912] & 0.114 [0.073, 0.142] & 0.666 [0.602, 0.705] & 0.829 [0.767, 0.890] \\
bland boilerplate   & 0.000 & 0.500 & 1.000 & 0.000 \\
\bottomrule
\end{tabular}
\caption{Synthetic validation: per-arm sequence-level summaries (median with central 95\% band over eight confirmatory sequences; the boilerplate control is deterministic). These validate instruments and arm machinery, not open-weight-model performance. The pure-sampling grid's $N$--$C$ correlation is $r=-0.735$.}
\label{tab:arms}
\end{table}

\subsection{Limits of the synthetic validation}
\label{sec:results:honest}

The synthetic sampling sweep shows a negative $N$--$C$ association ($r=-0.735$), and the mechanism has high novelty. Selection and sampling arms do not separate on $C_{\mathrm{sty}}$, however, and the accumulation curves do not distinguish mechanism from memory reset. The stand-in is strongly style-conditioned in every arm, producing a ceiling on the consistency metric. Consequently, these results validate the metric and code paths but do not evaluate C2 or C3 on a real model.

\begin{figure}[t]
\centering
\includegraphics[width=0.72\linewidth]{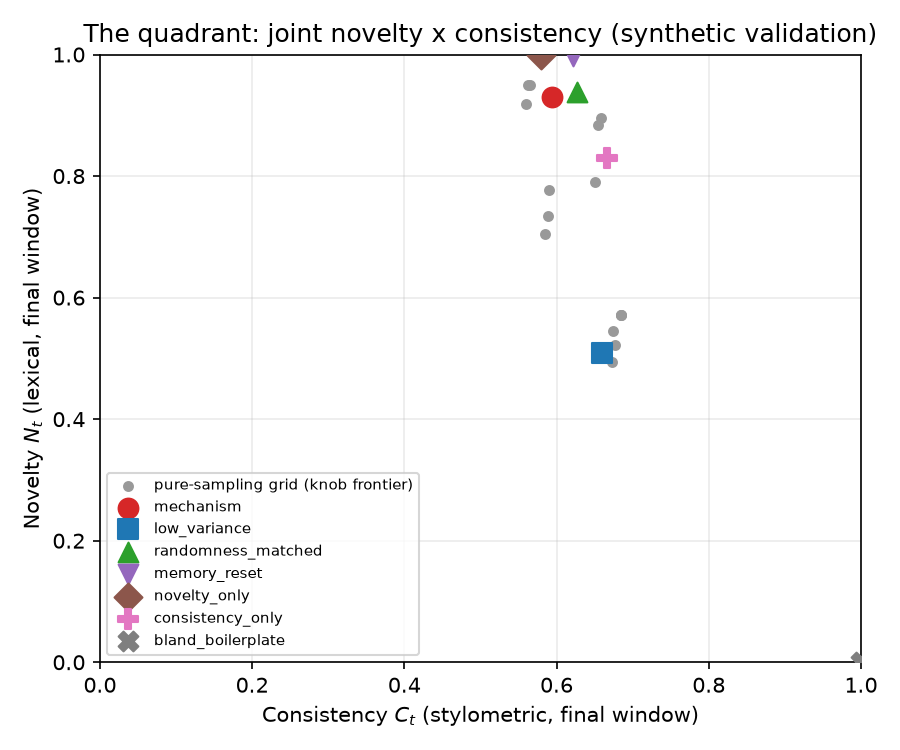}
\caption{Synthetic quadrant. Grey points are the pure-sampling temperature $\times$ nucleus-mass grid; markers are the six core arms and boilerplate control. The synthetic arms do not separate on $C_{\text{sty}}$; the real-model comparison was prospectively specified.}
\label{fig:quadrant}
\end{figure}

\subsection{The twin experiment: prediction versus measurement}
\label{sec:results:twins}

The divergence prediction was generated by the LLM-free state-space simulator and frozen---SHA-256 \texttt{dd7a5d81\ldots} recorded in the artifact, regeneration byte-identical, the reproduction script hard-failing on any drift---before any generation run. The frozen prediction: monotone fraction $0.972$; saturating curvature; half-to-full-horizon ratio $0.558$; floor-separation factor $3.25$.

\begin{figure}[t]
\centering
\includegraphics[width=0.75\linewidth]{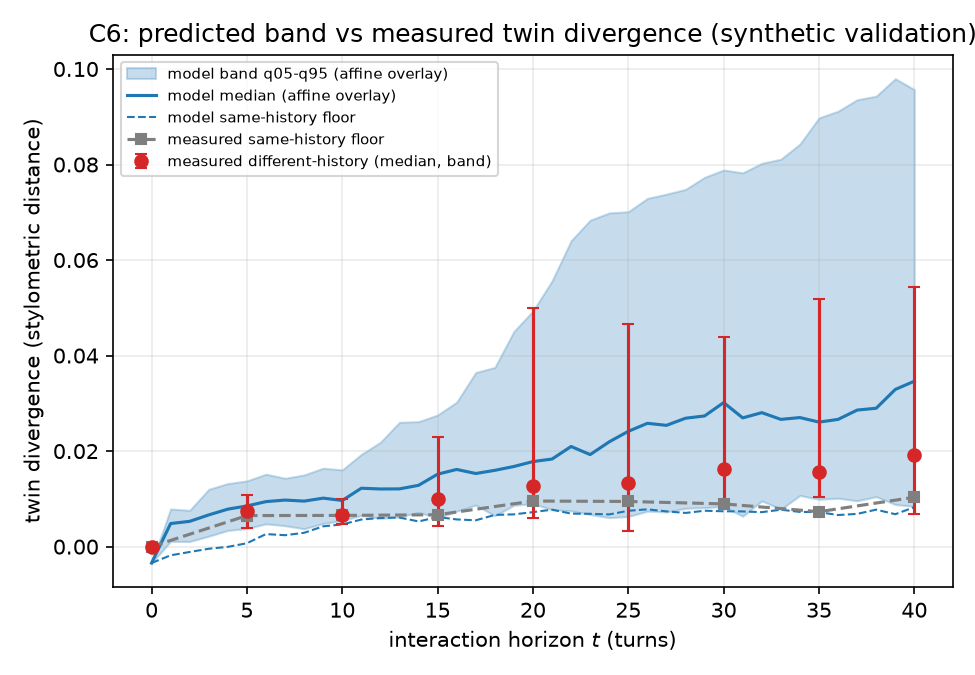}
\caption{Predicted band versus measured twin divergence, synthetic validation. Shaded: the frozen state-space band (5th--95th percentile) and its median, mapped through the pilot-fit affine overlay; dashed: the model's same-history floor. Points: measured different-history divergence (median with band over confirmatory pairs); squares: the measured same-history floor. The confirmatory shape comparison uses raw curves; the overlay is presentation only.}
\label{fig:divergence}
\end{figure}

The measured twins (twelve pairs per condition, 40 turns, two pilot pairs reserved for the figure's affine overlay, ten confirmatory) reproduce the predicted shape on all four point criteria (Table~\ref{tab:agreement}, Figure~\ref{fig:divergence}): the confirmatory divergence curve is monotone (fraction $1.000$), saturating, with a half-to-full ratio of $0.662$ and a floor-separation factor of $1.87$. Different-history twins diverge; same-history twins, differing only in state noise, diverge less by the pre-stated margin; the competitor classes predict a factor of one (Section~\ref{sec:model:twins}). Per the intervals-not-points commitment: final divergences $0.019\,[0.007,\,0.054]$ against $0.010\,[0.004,\,0.034]$, and the floor-factor bootstrap interval $[0.85,\,5.07]$ at ten pairs does not yet exclude one---the point criterion passes; the interval question passed to the real grids. The same compressive-transform argument applies to the floor factor's level: both measured features sit closer to one than their state-space counterparts ($1.87$ vs.\ $3.25$) without that being evidence against path dependence. In the synthetic setting, the model's central prediction---identity accumulated from selection makes initially identical agents diverge as a function of what happened to them, along a curve of the predicted shape---is corroborated \emph{at point level}: the discriminating feature's interval question stays open at ten pairs, under the same intervals discipline the real-model grid is later held to.

Three disclosures accompany the agreement verdict. First, the ratio tolerance was widened during validation from $\pm 0.15$ to $\pm 0.20$ once the shift's source was understood---the probe measures a \emph{bounded} profile distance, a compressive transform, so measured curves saturate slightly early; the widening is recorded in the protocol as a validation-stage calibration of a provisional constant, the widened tolerance still fails a flat or generic-memory process (ratio near one), and the measured ratio passes at the \emph{original} $\pm 0.15$ anyway. Second, the probe instrument was revised once during piloting---probes now condition on the agent's state alone (Section~\ref{sec:protocol:twins}); mixed conditioning had let the probe question pin the topic channel---and all numbers here are from the revised instrument, documented in the protocol. Third, the affine overlay in Figure~\ref{fig:divergence} (fit on the two pilot pairs) is presentation only; every number in Table~\ref{tab:agreement} is computed on raw curves.

\begin{table}[t]
\centering
\small
\begin{tabular}{@{}lcccc@{}}
\toprule
\textbf{Shape feature} & \textbf{Predicted (frozen)} & \textbf{Measured} & \textbf{Criterion} & \textbf{Pass} \\
\midrule
monotone fraction      & 0.972      & 1.000      & $\ge 0.80$        & yes \\
curvature class        & saturating & saturating & equal             & yes \\
half-to-full ratio     & 0.558      & 0.662      & within $\pm 0.20$ & yes \\
floor separation       & 3.25       & 1.87       & $\ge 1.5$         & yes (point) \\
\bottomrule
\end{tabular}
\caption{Synthetic twin-divergence agreement against prospectively hash-frozen shape features. All four point criteria pass, including the original $\pm0.15$ ratio tolerance, but the floor-separation interval at ten pairs includes one. Agreement is therefore point-level pipeline validation.}
\label{tab:agreement}
\end{table}

\subsection{A real-model pilot: the transmission risk, realized}
\label{sec:results:pilot}

The first open-weight run used \texttt{mlx-community/Qwen2.5-0.5B-Instruct-4bit}: six confirmatory 12-turn sequences per arm, four twin pairs per condition over 24 turns, real topic lexicons, and a fixed voice renderer. Pilot constants were frozen in the protocol. Calibration produced $\theta_N=0.95$ and $\theta_C=0.70$ under the floor-to-$0.05$ rule; the semantic threshold was $0.0$ and therefore uninformative. The frozen divergence band was evaluated on its pilot-horizon prefix without refitting.

The mechanism meets the informative thresholds ($N_{\text{lex}}=0.975\,[0.925,0.984]$, $C_{\text{sty}}=0.774\,[0.739,0.810]$), lies outside the tested sampling envelope, and has positive novelty contrasts: $+0.139\,[+0.091,+0.234]$ against the low-variance control and $+0.054\,[+0.008,+0.075]$ against consistency-only selection. Probe accuracy ranges from $0.85$ to $0.90$ with overlapping Wilson intervals, but $n=20$ per arm is insufficient for the planned equivalence test. Selection costs $598\,[560,702]$ microseconds per turn versus $1.56$ seconds for generation.

Both consistency contrasts are null. Against matched sampling, the difference is $-0.007$ (95\% CI $[-0.047,+0.032]$). Against novelty-only selection, it is $-0.014$ (95\% CI $[-0.057,+0.033]$). All arms lie near $C_{\mathrm{sty}}=0.77$--$0.80$, accumulation curves plateau within two turns, and neither twin channel exceeds the estimator floor. The results are consistent with weak transmission through the 0.5B model and minimal renderer: hidden states diverge in the wrapper, but measured function-word rates, sentence lengths, and topic mixtures do not. Before a powered grid, the protocol therefore required a renderer-level manipulation check on the measured channels. The protocol documents two pilot-stage instrument revisions---per-variant thresholds and state-only probe conditioning---made before the powered grid.

\subsection{The renderer gate: a manipulation check on the transmission channel}
\label{sec:results:gate}

The gate varies one block of $H$ at a time and asks whether the corresponding surface channel responds ($n{=}24$ generations per cell, fixed decoding). Three channels match what the instruments read: sentence length (high-vs-low gap of at least two words, bootstrap interval excluding zero), function-word group rates (one-hot a group; its rate must rise, two of three groups), and topic (argmax matches $H$'s dominant topic at $\ge 0.8$, measured through the pipeline's actual carrier, the question renderer). Two check-stage revisions are documented in the protocol: inflection-tolerant lexicon matching (``stars'' counts for ``star'') and the ecological topic measurement just described.

The 0.5B model fails the style channels (length gap $3.33$ words, interval $[-0.46,7.09]$; all function-word groups null) while passing topic ($0.90$). The question renderer largely carries the topic signal, whereas the model must realize style. \texttt{mlx-community/Qwen2.5-1.5B-Instruct-4bit} passes all three channel rules: length gap $5.74$ words $[3.35,8.13]$; positive effects for two of three function-word groups; and topic accuracy $0.88$. It was the smallest tested model to pass and was frozen for both grids.

\subsection{The powered grid: novelty without joint consistency}
\label{sec:results:powered}

Across the two grids, the response selector increases lexical novelty without a detected loss on the fixed capability probes. The joint criterion fails because consistency does not improve over sampling. The original C3 fails, while a later registered state-retention contrast is positive. Twin separation is not established; the final mean curve is classified as saturating. Table~\ref{tab:glance} separates these outcomes.

\begin{table}[t]
\centering
\small
\begin{tabular}{@{}lll@{}}
\toprule
\textbf{Claim} & \textbf{Evidence class} & \textbf{Real-model verdict} \\
\midrule
Formal model, entropy gap (Section~\ref{sec:model}) & argument & design target; not measured \\
Joint novelty$\times$consistency check (C2)   & measurement & \textbf{fail}: consistency null (one equiv.\ to zero) \\
Original accumulation criterion (C3)          & measurement & \textbf{fail}: arm-specific rise not observed \\
Final-grid state-retention contrast            & measurement & positive on $C$; not evidence of path dependence \\
Capability neutrality                         & measurement & established at probe level, item-paired \\
Twin path dependence and shape                & measurement & decided: separation not estab.; curvature as predicted \\
Handoff to the perception study               & conditional & withheld: manipulation-check condition not met \\
\bottomrule
\end{tabular}
\caption{Results at a glance: the paper's central claims, their evidence class, and the verdicts after the powered and final grids (Sections~\ref{sec:results:powered}--\ref{sec:handoff}).}
\label{tab:glance}
\end{table}

Powered-grid constants were frozen before execution: 24 confirmatory sequences per arm; eight 24-turn twin pairs per condition (two overlay pilots, six confirmatory); and $n=150$ probes per arm with a $\pm0.10$ equivalence margin. Calibration produced $\theta_N=0.95$, $\theta_C=0.70$, and an uninformative semantic threshold of $0.0$; novelty matching gave temperature $1.000$ (target $0.984$, achieved $0.972$). The registered permutation specification did not state the exchangeability group, and the run-time analysis permuted arm labels without preserving script pairs. After outcome access, this was corrected and labeled as a post-data reanalysis: medians of within-script differences were tested by script-level sign flips, with Holm correction. Both analyses give the same qualitative decisions; Table~\ref{tab:powered} reports the paired reanalysis. The powered-grid sizing calculation was an unadjusted normal approximation, not a Holm-aware guarantee. The final grid replaced it with simulation-based sizing.

Probe accuracy is $0.99$--$1.00$ in every arm, and the planned normal-approximation TOSTs pass at the $\pm0.10$ margin. Because the same near-ceiling items are shared across arms and item-level outcomes were not archived for this grid, the final grid provides the stronger paired analysis. Selection costs $604$ microseconds per turn versus $2.11$ seconds for generation; output quality was not assessed. Both lexical novelty contrasts pass in the paired reanalysis: $+0.092\,[+0.059,+0.111]$ against the low-variance control and $+0.021\,[+0.007,+0.033]$ against consistency-only selection, $p_{\mathrm{holm}}=0.0012$ for both. The semantic threshold of $0.0$ makes the registered two-variant novelty conjunct non-evaluable at this stage. Memory reset retains the same lexical novelty ($-0.008\,[-0.019,0.000]$ relative to mechanism), locating the effect in per-turn response selection rather than state accumulation. Twin divergence is suggestive at point level, but its paired interval includes zero; no path-dependent separation is established in this grid.

Two boundaries emerge. First, both consistency contrasts are null, so the joint criterion fails. The mechanism-minus-control estimates are $-0.019$ (95\% CI $[-0.030,+0.005]$) for matched sampling and $0.000$ (95\% CI $[-0.018,+0.026]$) for novelty-only selection; both Holm-adjusted $p$-values equal 1.0 (raw $0.996$ and $0.507$). These superiority tests do not themselves establish equivalence; that rule was added to the final grid. House-style dominance is one interpretation, but attenuation of the function-word metric or a weak realized state signal remain possible. The original accumulation criterion (C3) also fails: $C_t$ plateaus within roughly three turns in every arm, and the memory-reset ablation does not fall below the mechanism ($0.000\,[-0.016,+0.018]$, descriptive). The final grid evaluates a revised paired state-retention contrast, defined after this powered result and frozen before the final grid; it does not change the original C3 verdict. The renderer gate shows that the model can change style under direct one-hot instructions but does not separate the tested arms under independent-turn prompting. Second, the twin point estimate has accelerating rather than saturating curvature (half-to-full ratio $0.434$ versus $0.772\pm0.20$, the frozen band's ratio on the real runs' 24-turn checkpoint prefix; the full 40-turn synthetic horizon gives the $0.558$ of Table~\ref{tab:agreement}), but its pairs-bootstrap interval $[0.10,3.18]$ is too wide to establish either agreement or disagreement. The final grid supplies the registered interval-based twin decisions.

\begin{table}[t]
\centering
\small
\begin{tabular}{@{}llll@{}}
\toprule
\textbf{Specified outcome} & \textbf{Value} & \textbf{Criterion} & \textbf{Result} \\
\midrule
$N_{\text{lex}} \ge \theta_N$              & 0.984 [0.926, 1.000] & $\ge 0.95$ & pass \\
$C_{\text{sty}} \ge \theta_C$              & 0.745 [0.692, 0.814] & $\ge 0.70$ & pass \\
Outside sampling envelope                  & yes                  & required   & pass \\
$N$ vs.\ greedy floor                      & $+0.092$ [$+0.059$, $+0.111$] & $p_{\text{holm}}{=}0.0012$ & pass \\
$N$ vs.\ consistency-only                  & $+0.021$ [$+0.007$, $+0.033$] & $p_{\text{holm}}{=}0.0012$ & pass \\
$C$ vs.\ matched randomness                & $-0.019$ [$-0.030$, $+0.005$] & $p_{\text{holm}}{=}1.0$    & \textbf{fail} \\
$C$ vs.\ novelty-only                      & $+0.000$ [$-0.018$, $+0.026$] & $p_{\text{holm}}{=}1.0$    & \textbf{fail} \\
Joint manipulation check (conjunction)     & ---                  & all of the above & \textbf{fail} \\
Accum.: memory-reset collapse (descr.)     & $0.000$ [$-0.016$, $+0.018$] & collapse below mechanism & \textbf{fail} \\
Accum.: arm-specific $C_t$ rise (descr.)   & plateau by $\sim$turn 3, all arms & sustained, arm-specific & \textbf{fail} \\
TOST vs.\ greedy floor                     & $0.99$ vs.\ $1.00$   & $\pm 0.10$, $p<0.0001$ & equivalent \\
TOST vs.\ matched randomness               & $0.99$ vs.\ $0.99$   & $\pm 0.10$, $p<0.0001$ & equivalent \\
TOST vs.\ memory-reset (addendum)          & $0.99$ vs.\ $0.99$   & $\pm 0.10$, $p<0.0001$ & equivalent \\
Twin monotone fraction                     & 1.000                & $\ge 0.80$ & pass (point) \\
Twin floor separation (ratio)              & unbounded (zero denom.) & $\ge 1.5$  & trivial \\
Twin floor difference (post hoc)           & $+0.014$ [$0.00$, $+0.044$] & CI excludes 0 & \textbf{not met} \\
Twin curvature class                       & accelerating         & saturating & \textbf{fail} \\
Twin half-to-full ratio                    & 0.434                & $0.772 \pm 0.20$ & \textbf{fail} \\
\bottomrule
\end{tabular}
\caption{Powered grid, prospectively specified outcomes (\texttt{mlx-community/Qwen2.5-1.5B-Instruct-4bit}; 24 scripted sequences/arm; twins 8 pairs/condition, 2 overlay pilots, 6 confirmatory; capability $n{=}150$/arm). The memory-reset TOST is a post-review addendum using the same probes, margin, and seeds. Medians have central 95\% bands; contrast rows use the archived paired sign-flip reanalysis. Twin point criteria have no robust interval verdict at this size.}
\label{tab:powered}
\end{table}

\begin{figure}[t]
\centering
\includegraphics[width=0.7\linewidth]{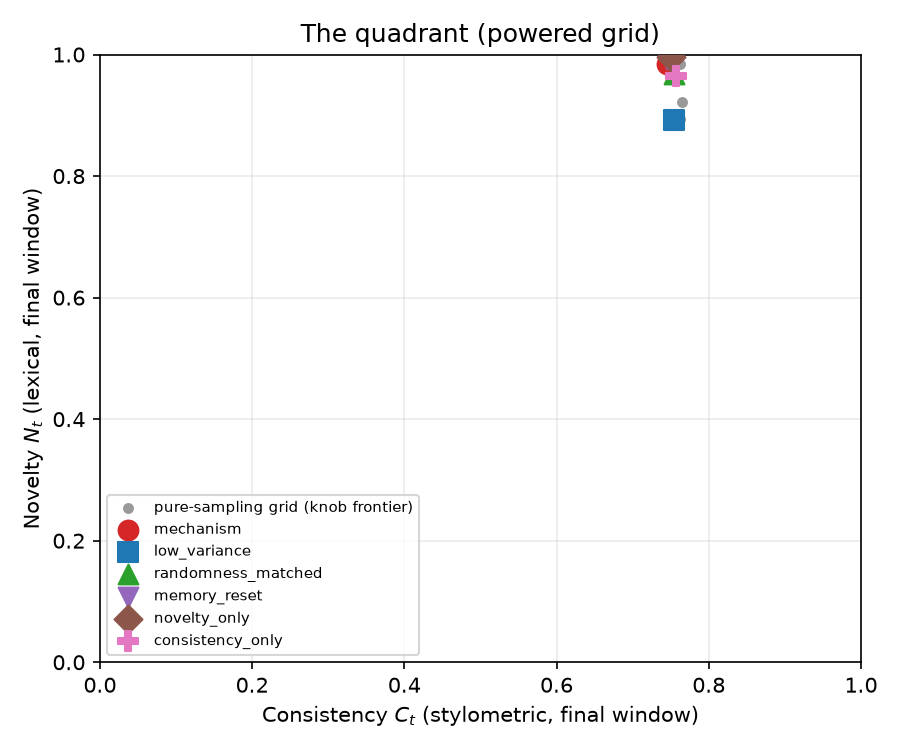}
\caption{Powered-grid quadrant (1.5B grid model, 24 scripted sequences/arm). The mechanism clears the frozen novelty threshold and tested sampling envelope; all arms remain within a narrow consistency band. Generated from \texttt{results\_powered/}.}
\label{fig:quadrant_powered}
\end{figure}

\begin{figure}[t]
\centering
\includegraphics[width=0.75\linewidth]{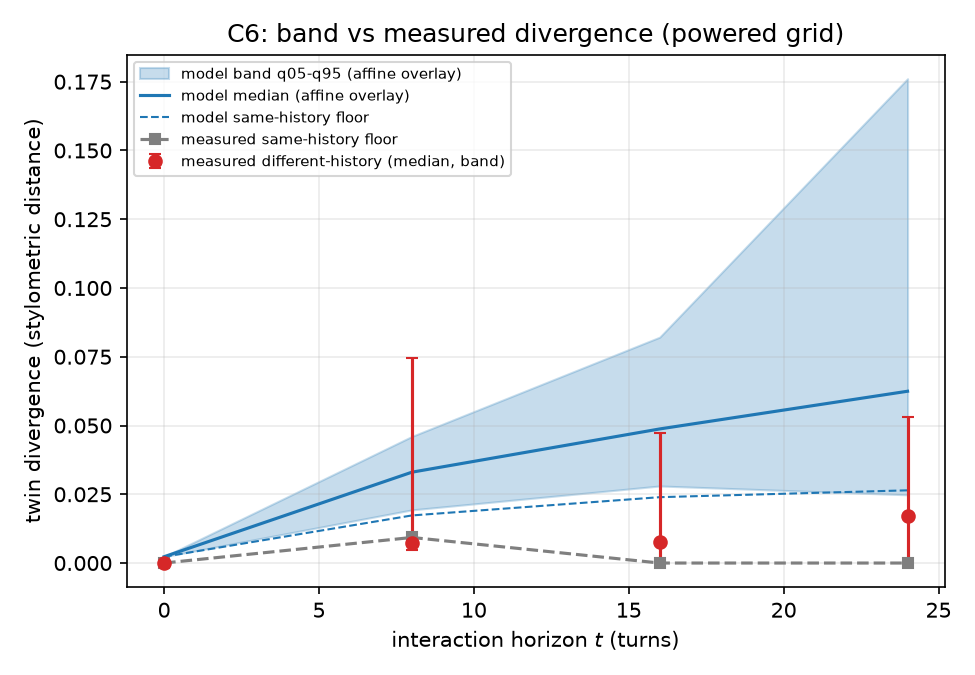}
\caption{Powered-grid twin divergence against the frozen horizon prefix (affine overlay for presentation only). The different-history point curve is monotone and accelerating, but intervals do not establish either above-floor separation or shape agreement. Final-grid results appear in Figure~\ref{fig:divergence_final}.}
\label{fig:divergence_powered}
\end{figure}

\subsection{The final grid: prospectively specified decisions}
\label{sec:results:final}

The final-grid design, sizing, and analysis script were frozen in repository commit \texttt{23c269a} at 20:03 EEST on 11 July 2026, before execution. Calibration thresholds and novelty matching were generated at 20:06--20:07. The OSF registration (\url{https://osf.io/dyt3w}) was created at 20:33:17 EEST, after those calibration artifacts but before confirmatory arm outputs began at 21:33. Thus the decision rules were prospectively fixed in version control, while the independent registration preceded confirmatory outcomes but not the entire run. Sizing used simulations based on powered-grid variances: at the chosen sizes---56 confirmatory scripted sequences per arm, 16 confirmatory twin pairs plus two overlay pilots, and $n{=}150$ capability items per arm, including memory reset---the archived simulations put the four contrasts at or above $94\%$ Holm-aware power at the planning effects, the equivalence declaration at $86\%$ under a true zero, and the twin separation test at $97\%$ under the powered grid's empirical zero-inflated effect model, which incorporates the estimator's clip. The final run used master seed 20260716, calibration thresholds $\theta_N=0.95$ and $\theta_C=0.65$ (produced by the same frozen floor-to-$0.05$ calibration rule as the earlier stages, applied to this grid's own calibration split), and matched temperature $1.400$ (target $0.990$, achieved $0.987$); on the confirmatory split the balance held---median $N_{\mathrm{lex}}$ $0.984$ for the mechanism versus $0.980$ for the matched arm, a difference of $0.003$ against the $0.03$ matching tolerance (descriptive, from the archived per-sequence files). Primary inference uses paired sign flips, paired bootstrap intervals, a semantic non-inferiority guard, and a mean paired twin statistic. The frozen analysis script generates all verdicts from \texttt{results\_final/}.

The lexical novelty contrasts are positive: $+0.073\,[+0.057,+0.103]$ against the low-variance control and $+0.023\,[+0.018,+0.034]$ against consistency-only selection, with $p_{\mathrm{holm}}=0.0004$ for both; both contrasts operate near the top of the bounded scale (mechanism median $0.984$). The absolute thresholds, sampling-envelope rule, and semantic non-inferiority guard also pass. In contrast, mechanism versus novelty-matched sampling on $C_{\mathrm{sty}}$ is $-0.006\,[-0.017,+0.008]$ (raw $p=0.71$, $p_{\mathrm{holm}}=0.92$) and meets the registered equivalence-to-zero rule; the novelty-only contrast (raw $p=0.46$) is null but does not meet that equivalence rule. The equivalence verdict is not sensitive to the margin's calibration: under the registered one-sided rule, the matched-sampling upper bound of $+0.008$ also satisfies margins of $0.01$ and $0.03$. The joint novelty--consistency criterion therefore fails.

The original C3 remains failed because the required arm-specific rise in $C_t$ is absent. The revised final-grid state-retention contrast is positive: mechanism exceeds memory reset on $C_{\mathrm{sty}}$ by $0.028\,[0.018,0.039]$ ($p=10^{-4}$), subject to the instrument caveat of Section~\ref{sec:boundary}: the stylometric metric's discriminant validity was not separately re-established on real-model outputs. Because this estimand was introduced after the powered-grid results, it is a prospectively specified follow-up within the final grid, not confirmation of the original C3. It also does not establish path dependence: a fixed nonzero control state could produce the same contrast.

\begin{figure}[t]
\centering
\includegraphics[width=0.75\linewidth]{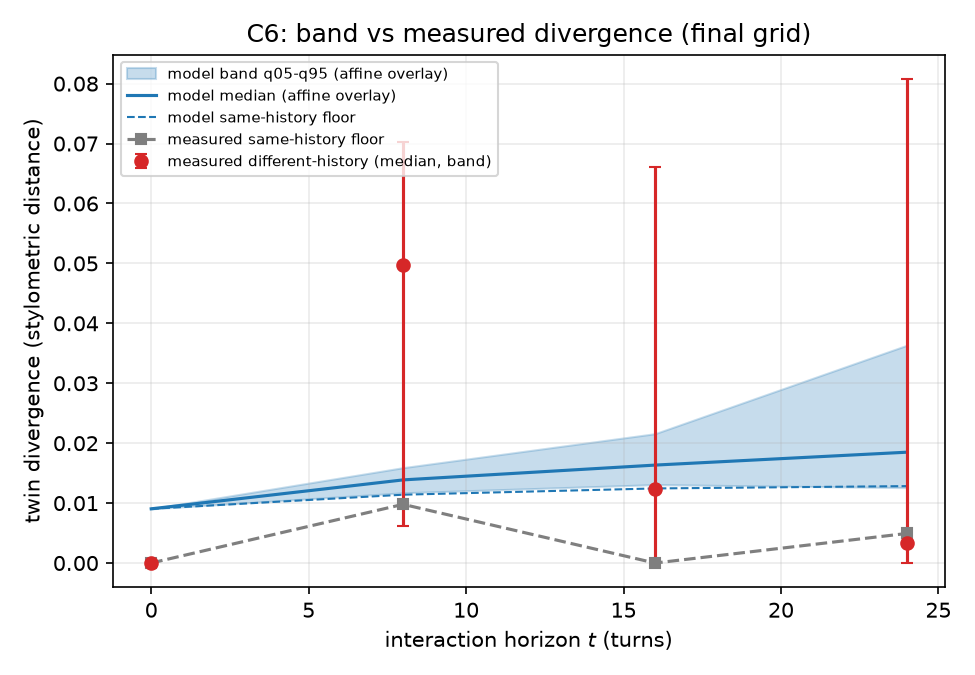}
\caption{Final-grid twin divergence vs.\ the frozen band's horizon prefix (affine overlay, presentation only; 16 confirmatory pairs). Points are the median with band over confirmatory pairs and are noisy at this scale; the registered verdicts (Section~\ref{sec:results:final}) are computed on the mean curve: above-floor separation not established and tightly bounded; curvature established saturating, as the drift model predicted. Generated from \texttt{results\_final/}.}
\label{fig:divergence_final}
\end{figure}

Twin separation is not established (mean paired difference $0.003\,[-0.011,0.019]$, $p=0.36$, 16 confirmatory pairs), the point estimate lying below the estimator's simulated null mean of ${\approx}0.006$; the superseded median statistic sits at the clip floor (median paired final difference $0.0$ over archived pairs, descriptive). The registered $97\%$ power applies to the powered grid's empirical planning effect ($0.039$ after the clip), not to effects at the estimator's floor: separations at the observed scale lie below the instrument's resolution. The mean curve is classified as saturating in $99.9\%$ of bootstrap resamples, but its half-to-full ratio $1.33\,[0.80,2.96]$ is not contained within the frozen band. The registered shape rules evaluate the mean curve, the median being fragile under the estimator's zero-clip; monotonicity had no registered mean-curve rule, and the median curve plotted in Figure~\ref{fig:divergence_final} is non-monotone at this scale (monotone fraction $0.333$) and carries no registered verdict. With no above-floor separation, curvature agreement alone does not support history-dependent divergence. Finally, the three item-paired capability comparisons have bootstrap intervals within $\pm0.10$ on the fixed 150-item battery (ten keyword-scored factual probes crossed with fifteen seeds; no exclusions; items paired across arms by probe and seed; arm accuracies $0.95$--$1.00$, near ceiling); the items bootstrap treats the probe$\times$seed cells as exchangeable, so the equivalence intervals are conditional on this fixed battery. The widest combined range is $[-0.047,+0.067]$. Selection overhead is $688$ microseconds per turn versus $1.81$ seconds for generation. Conversational output quality was not measured.

\begin{table}[t]
\centering
\small
\begin{tabular}{@{}llll@{}}
\toprule
\textbf{Registered outcome} & \textbf{Value} & \textbf{Criterion} & \textbf{Result} \\
\midrule
$N_{\text{lex}} \ge \theta_N$            & 0.984 [0.959, 1.000] & $\ge 0.95$ & pass \\
$C_{\text{sty}} \ge \theta_C$            & 0.764 [0.689, 0.825] & $\ge 0.65$ & pass \\
Outside sampling envelope                & yes & required & pass \\
Semantic guard (non-inferiority)         & $-0.004$ [$-0.016$, $0.000$] & lower bound $> -0.05$ & holds \\
$N$ vs.\ greedy floor                    & $+0.073$ [$+0.057$, $+0.103$] & $p_{\text{holm}}{=}0.0004$ & pass \\
$N$ vs.\ consistency-only                & $+0.023$ [$+0.018$, $+0.034$] & $p_{\text{holm}}{=}0.0004$ & pass \\
$C$ vs.\ matched randomness              & $-0.006$ [$-0.017$, $+0.008$] & $p_{\text{holm}}{=}0.92$ & \textbf{fail}; equiv.\ to 0 \\
$C$ vs.\ novelty-only                    & $+0.001$ [$-0.009$, $+0.021$] & $p_{\text{holm}}{=}0.92$ & \textbf{fail} \\
Joint manipulation check                 & --- & all conjuncts & \textbf{fail} \\
Original C3 (two-part criterion)         & no arm-specific rise & both parts required & \textbf{fail} \\
Final-grid reset contrast ($C$)          & $+0.028$ [$+0.018$, $+0.039$] & $p{<}.05$, CI $>0$ & positive \\
Twin separation (mean paired diff)       & $+0.003$ [$-0.011$, $+0.019$] & $p{<}.05$, CI $>0$ & \textbf{not estab.} \\
Twin curvature (mean curve)              & saturating (99.9\% resamples) & $=$ predicted & \textbf{established} \\
Twin half-to-full ratio (mean curve)     & 1.33 [0.80, 2.96] & CI inside $0.772 \pm 0.20$ & not estab. \\
Capability (item-paired, 3 pairs)        & all within $[-0.047, +0.067]$ & CIs inside $\pm 0.10$ & equivalent \\
\bottomrule
\end{tabular}
\caption{Final-grid outcomes (\texttt{mlx-community/Qwen2.5-1.5B-Instruct-4bit}; 56 scripted sequences/arm; twins 18 pairs/condition, 16 confirmatory; capability $n{=}150$/arm including memory reset). Paired medians, except the twin mean, have pairs-bootstrap 95\% intervals; sign-flip tests are Holm-corrected where specified. Repository rules were frozen before execution; OSF registration followed calibration and preceded confirmatory outputs.}
\label{tab:final}
\end{table}

\section{The Ablation, Packaged for the Perception Study}
\label{sec:handoff}

The framework paper requires a capability-neutral present-versus-ablated condition pair before the perception study~\cite{cochinescu2026perceivedagi}. The candidate pair is the full mechanism versus memory reset, with the joint criterion as its manipulation check and twin separation as an independent path-dependence signature.

The handoff condition was not met. Probe-level capability equivalence holds for the mechanism--memory-reset pair, but the joint manipulation check fails on both consistency contrasts. The positive final-grid reset contrast shows that the two arms are behaviorally distinguishable on $C_{\mathrm{sty}}$, subject to the instrument caveat of Section~\ref{sec:boundary}; it does not make the full arm an entropy-present condition because the registered joint definition also requires superiority to matched sampling and the novelty-only selector. The leading interpretation is a narrow house-style band under the tested independent-turn renderer, with metric attenuation and weak realized state signal remaining alternatives. A human study would still require its own manipulation check, output-quality validation, participant sample, perceived-mind measure, and disclosure conditions. No machine-judge result is used as a substitute for human perception.

\section{Scope and Limitations}
\label{sec:boundary}

\textbf{Inferrable interiority, not interiority.} The object of this paper is the evidence structure from which an observer could infer a hidden state---unpredictable-yet-coherent outputs---not the possession of one in any phenomenal sense. This is perception engineering, not a theory of machine consciousness; the register is the framework paper's~\cite{cochinescu2026perceivedagi}, consonant with reading an LLM's apparent inner states as role-play rather than report~\cite{shanahan2023roleplay}. Nothing here claims the mechanism increases perceived mind, believability, or any human rating: those are the perception study's questions, and the bridge---does the high/high quadrant move perceived mind---is the framework's prediction P2, which this paper's deliverable makes pre-registrable without testing it. The title names the design target, not the real-model result: on real models the evidence reaches the novelty channel, at the selector level, and the paper claims nothing further.

\textbf{No human subjects; independent-turn evaluation.} Every result is machine-scored on one computer. The synthetic generator was designed to transmit conditioning, so its positive twin result validates the pipeline but does not predict open-weight-model performance. The real-model renderer gate established large one-hot style effects, yet arm-level consistency remained within a narrow band. Most importantly, the base model never receives the preceding dialogue: each call contains only the current rendered state and prompt. The experiment therefore isolates wrapper state and selection under a scripted sequence; it is not evidence about long-context dialogue, memory retrieval, or persona drift in sustained interaction. The ambient stream is also simulated. Because stream items and current-turn features enter the same state update, their separate contributions are not identified.

\textbf{Proxy outcomes.} $N_t$ and $C_t$ measure non-repetition and stylometric stability, not the entropy gap or a human perception. ``Identity'' denotes a selection-history construct, not a psychological property. Scripts, thresholds, metric definitions, per-sequence and per-item outcome files, and deterministic analysis code are included in the submission ancillary files and archived independently at \url{https://doi.org/10.5281/zenodo.21462383}; per-turn response text is not retained, so the topic-permutation and distinctiveness checks ran in-pipeline at the synthetic stage and were not recomputed on the real grids---the stylometric instrument's discriminant validity on real outputs rests on the renderer gate. The operative novelty result is lexical and could reflect surface substitution rather than substantive variation. The novelty metric and the response selector's novelty objective are aligned by construction, so the positive lexical contrasts show the selector operating as designed rather than providing independent evidence of substantive unpredictability. The final-grid semantic non-inferiority guard rules out a specified decrement relative to matched sampling; it is not evidence of positive semantic novelty.

\textbf{The controls limit interpretation.} ``More consistent than randomness'' refers to novelty-matched sampling; ``noveler than the floor'' refers to the fixed low-variance control. The original C3 means an arm-specific rise in consistency plus loss under reset and is not supported. The later mechanism--reset contrast is evidence that retained state affects $C_{\mathrm{sty}}$, not that the state follows a path-dependent trajectory. The twin comparison was intended to test that stronger claim and did not establish separation. Capability equivalence applies only to the fixed probes and measured latency; output quality was not evaluated.

\textbf{Model and software scope.} The powered and final grids use the 4-bit MLX conversion of Qwen2.5-1.5B-Instruct from the \texttt{mlx-community} repository. Its Hugging Face revision is \texttt{8b403126fc14f14cfc99b}\allowbreak\texttt{b4cfa72ecbc129ea677}; inference uses \texttt{mlx-lm} 0.31.3 and \texttt{mlx} 0.32.0. The artifact records seeds, scripts, thresholds, per-sequence and per-item outcome files, and analysis code. This single model, quantization, renderer, and software stack do not support claims about other model families or deployment settings.

\section{Series Context}
\label{sec:series}

This is Paper 2 of a series whose framework defines time, truth, entropy, and love as candidate behavioral stances for perceived mind~\cite{cochinescu2026perceivedagi}. The time paper supplies the open-loop oscillator used here~\cite{cochinescu2025scnllm}; companion manuscripts address truth and love with distinct mechanisms and evidence. The present results support only a lexical-novelty effect and a post-powered state-retention contrast under the tested wrapper. They do not establish the joint entropy condition, history-dependent twin divergence, output-quality neutrality, or a perceptual effect. The condition pair required by the later human study therefore remains unavailable.

\section*{Editorial assistance disclosure}

OpenAI Codex was used for editorial review and LaTeX revision. It did not generate experimental data or run the reported analyses. The author takes responsibility for the manuscript.

\bibliographystyle{plain}
\bibliography{Entropy}

\end{document}